\documentclass[letterpaper,10pt,conference]{IEEEtran}

\IEEEoverridecommandlockouts

\usepackage[T1]{fontenc}
\usepackage{amsmath}
\usepackage{booktabs}
\usepackage{graphicx}
\usepackage{multirow}
\usepackage{url}
\usepackage[hidelinks]{hyperref}
\usepackage{xcolor}
\usepackage{pgfplots}
\pgfplotsset{compat=1.18}
\hypersetup{pdftitle={Traceable World State: A Provenance-Aware State Representation and Deterministic Replay Framework for Robotic Systems},pdfauthor={Zoe Li},pdfsubject={Research preprint}}
\newcommand{\tws}{\textsc{TWS}}

\title{Traceable World State: A Provenance-Aware State Representation and Deterministic Replay Framework for Robotic Systems}

\author{
\IEEEauthorblockN{Zoe Li}
\IEEEauthorblockA{Independent Researcher\\
Seattle, WA, USA\\
\texttt{zoeli4@siggraph.org}}
}

\begin{document}

\maketitle
\thispagestyle{plain}

\begin{abstract}
Robotic systems operating over extended tasks must maintain a world
state assembled from observations that arrive at different times,
carry different confidence levels, and may later be revised.
Conventional scene representations commonly emphasize the latest
estimated state, making it difficult to determine where a fact came
from, reproduce an earlier decision context, or detect corruption in
a recorded execution.

This paper presents Traceable World State (\tws), a middleware-neutral
semantic representation and reference runtime for provenance-aware
robot world state. A \tws snapshot represents entities, relations,
observations, confidence, and revision metadata. Validated update
operations transform snapshots without mutating their inputs.
Ordered updates can be replayed deterministically, while a canonical
SHA-256 hash chain makes stored event logs tamper-evident.

We evaluate the prototype through schema and semantic conformance
tests, complete manipulation-state lifecycles, deterministic replay,
and targeted fault injection. The current implementation passes 38
tests on Python 3.10, 3.12, and 3.14. The evaluation detects corrupted
updates, broken hash links, sequence discontinuities, world
mismatches, revision discontinuities, and malformed records.
The results support \tws as a compact foundation for reproducible
robotic state-management experiments, while leaving
middleware integration, multi-writer coordination, and trusted
external checkpoints to future work.
Across a deterministic sample of ten public BEHAVIOR-1K task definitions, \tws imported 153 entities and 146 relations, and all generated snapshots passed runtime and command-line validation. On 103 public NVIDIA Unitree G1 simulated trajectories containing 78,369 source frames, \tws achieved exact terminal-state replay in 103/103 episodes and detected all 412 injected record corruptions with 1.72\% storage overhead over Plain JSONL.
\end{abstract}

\begin{IEEEkeywords}
robotic systems, semantic state, provenance, middleware-neutral frameworks,
deterministic replay, event logs, reproducibility
\end{IEEEkeywords}

\section{Introduction}

A robot rarely receives a complete and perfectly consistent
description of its environment. Instead, its world state is assembled
incrementally from cameras, force sensors, learned perception models,
task interfaces, and prior state. Observations may be delayed,
uncertain, contradicted, or reinterpreted. These conditions are
particularly important in long-horizon manipulation, where a failure
may occur many state transitions after the observation that caused
it.

Many robot systems expose a convenient current-state interface, but
the latest state alone is insufficient for answering questions such
as: Which observation supported this relation? What confidence was
assigned when a planner selected an action? Which update changed an
object identity? Can the state used in a failed run be reproduced?
Was a stored execution log modified after collection?

Existing knowledge-processing systems provide rich semantic
reasoning~\cite{tenorth2009knowrob,beetz2018knowrob2}, while semantic
mapping and scene-graph research structures spatial and object-level
knowledge~\cite{nuechter2008semantic,armeni20193dscenegraph}.
Robot middleware provides communication and execution
infrastructure~\cite{macenski2022ros2}. These capabilities are
valuable, but they do not by themselves define a compact,
middleware-independent contract for traceable state transitions and
verifiable replay.

We introduce Traceable World State (\tws), a semantic data model and
reference runtime designed around explicit provenance, immutable
updates, revision continuity, and deterministic replay. \tws is not a
planner, perception system, or control framework. It is a boundary
between these components: producers publish evidence-backed updates,
and consumers receive validated snapshots with an auditable history.

The scope is a middleware-neutral state-management framework and reference runtime. Humanoid tasks provide application cases rather than a hardware requirement. The evaluation tests representation, transition semantics, replay integrity, and computational overhead; it does not establish perception accuracy or closed-loop task performance.

The contribution is the state-update contract that connects evidence references, semantic invariants, revision continuity, and replay. JSON Schema validation and SHA-256 hashing are established components used to implement that contract, rather than new algorithms introduced here.

The contributions of this work are:

\begin{itemize}
    \item A versioned semantic representation for entities, relations,
    observations, confidence, provenance, and world revisions.
    \item A deterministic update engine with explicit invariants for
    identity, references, operation payloads, and revision continuity.
    \item A tamper-evident JSON Lines event-log format based on
    canonical serialization and a SHA-256 hash chain.
    \item A reproducible conformance and fault-injection evaluation
    covering complete state lifecycles and invalid-log conditions.
\end{itemize}

The implementation intentionally avoids dependencies on ROS, a
specific simulator, or robot hardware. This permits its use as a
portable experiment artifact and as a future interoperability layer
for humanoid systems.

\section{Related Work}

\subsection{Robot Knowledge and Semantic World Models}

KnowRob introduced knowledge processing for autonomous
robots~\cite{tenorth2009knowrob} and was later extended to combine
symbolic and data-driven reasoning~\cite{beetz2018knowrob2}.
RoboSherlock organizes perception as an explainable knowledge-enabled
process~\cite{beetz2015robosherlock}, while OpenEASE supports the
analysis and reuse of robot experience data~\cite{beetz2015openease}.
These systems demonstrate the value of structured robot knowledge.

Semantic mapping associates geometric environments with meaningful
objects and concepts~\cite{nuechter2008semantic}. Scene graphs extend
this organization by representing objects, spaces, and their
relations~\cite{armeni20193dscenegraph}. \tws adopts entities and
relations as basic representational elements, but focuses on the
runtime contract through which state changes are validated, attributed
to evidence, persisted, and replayed.

\subsection{Provenance and Reproducibility}

The W3C PROV data model formalizes relationships among entities,
activities, and agents~\cite{moreau2013prov}. \tws uses a narrower
robotics-oriented provenance mechanism: state items contain source
references to timestamped observations. This deliberate restriction
keeps runtime documents small while retaining the evidence path needed
to inspect state assertions.

Reproducibility in robotics requires more than retaining sensor data.
The ordering and interpretation of state changes also matter.
Distributed-system work has long established the importance of event
ordering~\cite{lamport1978time}. \tws therefore records an explicit
base revision for each update and rejects discontinuous sequences.

\subsection{Integrity of Recorded State}

Hash-linked structures can expose modification of ordered
records~\cite{merkle1987digital}. \tws applies this principle to robot
world-state updates. Unlike a digital signature, the current hash
chain does not establish authorship and cannot prevent complete log
replacement. Its purpose is narrower: to make changes within a
retained execution record detectable when a trusted final hash or
checkpoint is available.

\section{Design Requirements}

The design follows six requirements derived from long-horizon robot
experiments.

\begin{enumerate}
    \item \textbf{Explicit uncertainty:} confidence must be represented
    numerically instead of hidden in component-specific state.
    \item \textbf{Traceable evidence:} entities and relations may
    reference the observations that support them.
    \item \textbf{Temporal clarity:} observation time and update time
    must remain explicit.
    \item \textbf{Reference integrity:} relations must not point to
    missing entities, and source references must not point to missing
    observations.
    \item \textbf{Deterministic transition semantics:} the same valid
    snapshot and ordered updates must produce the same final snapshot.
    \item \textbf{Middleware independence:} the core representation
    must be usable without requiring a particular robot communication
    framework.
\end{enumerate}

The principal non-goals of version 0.1 are motion planning, task
planning, sensor synchronization, geometric estimation, access
control, and multi-writer coordination.

\section{Traceable World State Model}

\subsection{Snapshots}

A snapshot is a complete validated state at one revision. Its envelope
contains a schema version, world identifier, current revision,
previous revision, observation time, entities, relations, and
observations. Table~\ref{tab:model} summarizes the major elements.

\begin{table}[t]
\caption{Core elements of a \tws snapshot.}
\label{tab:model}
\centering
\small
\begin{tabular}{p{0.22\columnwidth}p{0.68\columnwidth}}
\toprule
Element & Purpose \\
\midrule
World ID & Separates independent state histories. \\
Revision & Orders validated snapshots and rejects stale updates. \\
Entity & Represents a physical, virtual, or conceptual object. \\
Relation & Represents a typed directed edge between two entities. \\
Observation & Records timestamped evidence from a sensor or source. \\
Source & Connects an entity or relation to supporting evidence. \\
Confidence & Encodes uncertainty in the closed interval $[0,1]$. \\
\bottomrule
\end{tabular}
\end{table}

Identifiers are stable strings within a world. Entity, relation, and
observation identifiers must each be unique. A relation contains
subject and object identifiers that must resolve to existing entities.
A source reference containing an observation identifier must resolve
to an observation in the same snapshot.

The representation distinguishes absence from uncertainty. A missing
optional property means that the property has not been asserted.
Confidence describes the strength of an assertion and does not turn a
missing value into a known value.

\subsection{Updates}

A \tws update contains a schema version, update identifier, world
identifier, base revision, operation, observation time, confidence,
provenance, and exactly one operation-specific payload. Version 0.1
supports four operations:

\begin{itemize}
    \item \texttt{upsert\_entity},
    \item \texttt{remove\_entity},
    \item \texttt{upsert\_relation}, and
    \item \texttt{remove\_relation}.
\end{itemize}

An upsert replaces an existing item with the same identifier or
appends a new item. Removal requires the target to exist. An entity
cannot be removed while a relation still references it. A relation
cannot be created unless both endpoint entities exist.

Given snapshot $S_r$ at revision $r$ and update $u$, the transition
function is

\begin{equation}
S_{r+1} = F(S_r,u),
\end{equation}

subject to

\begin{equation}
u.\mathrm{world\_id}=S_r.\mathrm{world\_id}
\end{equation}

and

\begin{equation}
u.\mathrm{base\_revision}=r.
\end{equation}

The resulting snapshot records $r$ as its previous revision and
$r+1$ as its new revision.

\subsection{Structural and Semantic Validation}

Documents are structurally checked against JSON Schema Draft
2020-12~\cite{jsonschema2020}. Structural validation covers required
fields, data types, enumerations, formats, value ranges, and
operation-specific payload shapes.

Semantic validation covers invariants that cross document locations:
identifier uniqueness, relation endpoint resolution, and observation
source resolution. The runtime validates the input snapshot, update,
and resulting snapshot. Invalid transitions do not return a partially
updated state.

\section{Runtime Architecture}

\begin{figure}[t]
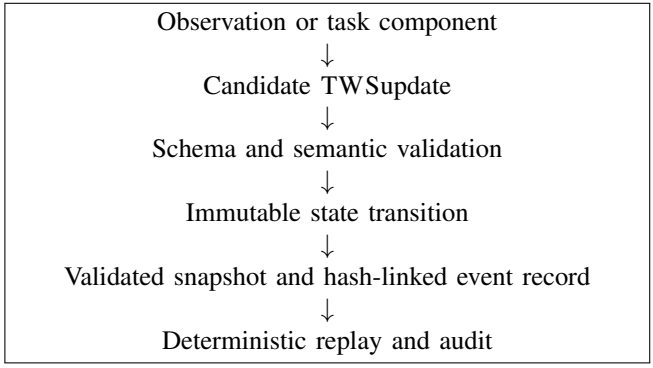

\centering
\fbox{
\begin{minipage}{0.91\columnwidth}
\centering
Observation or task component\\
$\downarrow$\\
Candidate \tws update\\
$\downarrow$\\
Schema and semantic validation\\
$\downarrow$\\
Immutable state transition\\
$\downarrow$\\
Validated snapshot and hash-linked event record\\
$\downarrow$\\
Deterministic replay and audit
\end{minipage}
}
\caption{Reference \tws runtime data flow.}
\label{fig:architecture}
\end{figure}

Figure~\ref{fig:architecture} shows the reference data flow. A
perception or task component emits a candidate update. The validator
checks its structure, after which the update engine checks the world,
revision, operation, and reference invariants. The resulting snapshot
is validated again. Accepted updates may then be stored as hash-linked
event records and replayed later.

The implementation deep-copies the input snapshot before applying an
operation. This provides a simple immutability guarantee at the API
boundary: a failed or successful update does not alter the caller's
input object.

Replay begins from a validated initial snapshot and applies updates in
stored order. Validation is repeated during replay. Consequently, a
log that is internally continuous but begins from the wrong world or
revision is rejected when combined with the initial snapshot.

\section{Tamper-Evident Event Log}

\subsection{Record Format}

Each event-log line is a UTF-8 JSON object containing

\begin{itemize}
    \item record format version,
    \item zero-based sequence number,
    \item previous record hash or null,
    \item one validated \tws update, and
    \item the current record hash.
\end{itemize}

Let $R_i$ denote record $i$. Its digest is

\begin{equation}
h_i =
\mathrm{SHA256}\left(
C(v_i,i,h_{i-1},u_i)
\right),
\end{equation}

where $C$ is canonical JSON serialization, $v_i$ is the record
version, and $u_i$ is the nested update. The stored hash is encoded as
the prefix \texttt{sha256:} followed by 64 lowercase hexadecimal
characters.

Canonical serialization sorts object keys lexicographically, removes
insignificant whitespace, preserves Unicode, encodes text as UTF-8,
and rejects non-finite numbers. The current record's hash field is
excluded from its own hash input.

\subsection{Append and Verification}

Before appending, a writer validates the complete existing chain and
the incoming update. It then calculates the next sequence number and
previous-hash link, serializes one record, flushes the file, and calls
the operating system synchronization operation. A rejected append
leaves the prior log unchanged.

A reader verifies every record before returning any updates. It stops
at the first invalid line and reports the one-based line number and
cause. Verification checks JSON syntax, record schema, nested update
schema, record sequence, hash linkage, calculated hash, world
continuity, and base-revision continuity.

\subsection{Security Boundary}

The log is tamper-evident rather than tamper-proof. If an attacker can
replace the full log and all external checkpoints, the attacker can
calculate a new internally consistent chain. A trusted final digest,
digital signature, remote timestamp, or independently retained
checkpoint is required to detect complete replacement.

The version 0.1 prototype also assumes a single writer. File locking,
distributed consensus, authorization, encryption, and crash recovery
from an interrupted final write are outside its present scope.

\section{Evaluation}

\subsection{Research Questions}

The evaluation addresses four questions:

\begin{itemize}
    \item \textbf{RQ1:} Does the implementation accept valid state
    lifecycles and reject invalid state transitions?
    \item \textbf{RQ2:} Does replay reproduce the expected final state
    without mutating the initial snapshot?
    \item \textbf{RQ3:} Does event-log verification detect targeted
    corruption and continuity faults?
    \item \textbf{RQ4:} What validation, append, and replay overhead is
    introduced as the number of updates increases?
\end{itemize}

\subsection{Experimental Setup}

The reference runtime is implemented in Python and uses the
\texttt{jsonschema} package for Draft 2020-12 validation. Continuous
integration evaluates Python 3.10, 3.12, and 3.14 on Linux. The
artifact contains one initial world snapshot and four ordered updates:
entity insertion, relation insertion, relation removal, and entity
removal.

The lifecycle represents a minimal manipulation-state sequence. An
object enters the tracked world, participates in a relation required
by a task, and is removed only after the relation has been removed.
Although deliberately small, this sequence exercises all four version
0.1 operations and the reference-integrity rule that prevents removal
of a referenced entity.

\subsection{Conformance and Lifecycle Results}

The current suite contains 38 tests. All tests pass under each of the
three Python versions in continuous integration. The tests cover
schema validation, semantic references, update preconditions,
immutability, ordered replay, event-log storage, command-line
behavior, packaging of schemas, and integrity checks.

For the complete four-update lifecycle, replay advances the snapshot
from revision 0 to revision 4 and records revision 3 as the immediate
predecessor. The original revision-0 snapshot remains unchanged.

\begin{table}[t]
\caption{Verified lifecycle and transition properties.}
\label{tab:conformance}
\centering
\small
\begin{tabular}{p{0.57\columnwidth}p{0.25\columnwidth}}
\toprule
Property & Result \\
\midrule
Valid four-operation lifecycle & Pass \\
Deterministic ordered replay & Pass \\
Input snapshot remains unchanged & Pass \\
Stale update rejection & Pass \\
Missing relation endpoint rejection & Pass \\
Referenced entity removal rejection & Pass \\
Schema-valid output snapshot & Pass \\
Python 3.10, 3.12, and 3.14 & Pass \\
\bottomrule
\end{tabular}
\end{table}

\subsection{Fault-Injection Results}

Fault injection modifies one property at a time in an otherwise valid
log. Table~\ref{tab:faults} reports the currently implemented cases.
Every injected fault is rejected before updates are returned for
replay.

\begin{table}[t]
\caption{Targeted event-log fault injection.}
\label{tab:faults}
\centering
\small
\begin{tabular}{p{0.55\columnwidth}p{0.27\columnwidth}}
\toprule
Injected fault & Detection result \\
\midrule
Modified nested update & Detected \\
Modified previous-hash value & Detected \\
Modified sequence number & Detected \\
World identifier mismatch & Detected \\
Base-revision discontinuity & Detected \\
Malformed JSON record & Detected \\
Non-object JSON record & Detected \\
Legacy unwrapped update record & Detected \\
Missing event-log file & Detected \\
\bottomrule
\end{tabular}
\end{table}

These results establish conformance for the tested cases, but they do
not prove resistance to arbitrary corruption or malicious input.

\subsection{Runtime Overhead}

We measured validation, append, complete-log verification, and
in-memory replay using event logs containing 10, 100, and 500 updates.
Each condition was executed five times after a warm-up run. Time was
measured using a monotonic high-resolution clock. We report the median
and interquartile range (IQR).

The experiment used CPython 3.14.4 on 64-bit Linux 7.0.0-28 with an
Intel Core Ultra 9 275HX processor and 24 logical CPUs. The benchmark
was single-process and did not control unrelated operating-system
activity.

\begin{table}[t]
\caption{Runtime results in milliseconds, reported as median (IQR)
over five repetitions.}
\label{tab:performance}
\centering
\small
\resizebox{\columnwidth}{!}{
\begin{tabular}{rrrrr}
\toprule
Updates & Validation & Append & Verification & Replay \\
\midrule
10  & 1.297 (0.029)  & 24.478 (1.207)
    & 2.944 (0.067)  & 9.197 (0.115) \\
100 & 12.588 (0.169) & 2111.450 (9.887)
    & 28.260 (0.124) & 91.053 (2.254) \\
500 & 60.883 (0.573) & 51941.915 (97.959)
    & 136.578 (1.748) & 438.086 (4.539) \\
\bottomrule
\end{tabular}
}
\end{table}

Validation, complete-log verification, and replay scaled approximately
linearly over the evaluated range. At 500 updates, validation required
60.883 ms, complete verification required 136.578 ms, and replay
required 438.086 ms. Their respective median costs were approximately
0.122, 0.273, and 0.876 ms per update.

Event-log size was 6,426 bytes for 10 updates, 63,154 bytes for 100
updates, and 316,954 bytes for 500 updates, corresponding to
approximately 634 bytes per stored record at the largest condition.

Append time increased superlinearly, reaching 51.942 s for 500
updates. The version 0.1 writer verifies the complete existing chain
before every append. Consequently, constructing a log through repeated
single-record appends performs cumulative work that is approximately
quadratic in the number of records. This conservative behavior
prioritizes integrity and leaves the previous log unchanged after a
rejected append, but it is not suitable for high-rate state streams.
A persistent verified writer, checkpointed verification, or batched
append interface is required before deployment at task-runtime event
rates.

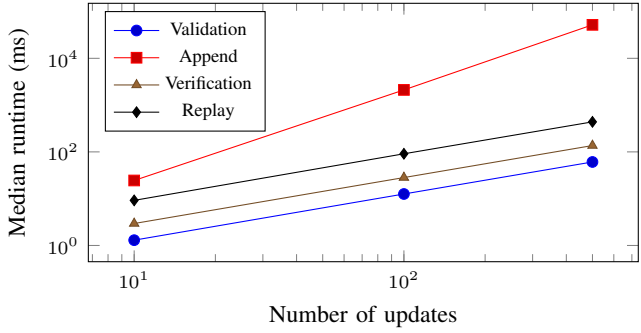
\begin{figure}[t]
\centering
\begin{tikzpicture}
\begin{loglogaxis}[width=\columnwidth,height=5cm,xlabel={Number of updates},ylabel={Median runtime (ms)},legend style={font=\scriptsize,at={(0.03,0.97)},anchor=north west},tick label style={font=\scriptsize},label style={font=\small}]
\addplot+[mark=*] coordinates {(10,1.297)(100,12.588)(500,60.883)};
\addlegendentry{Validation}
\addplot+[mark=square*] coordinates {(10,24.478)(100,2111.450)(500,51941.915)};
\addlegendentry{Append}
\addplot+[mark=triangle*] coordinates {(10,2.944)(100,28.260)(500,136.578)};
\addlegendentry{Verification}
\addplot+[mark=diamond*] coordinates {(10,9.197)(100,91.053)(500,438.086)};
\addlegendentry{Replay}
\end{loglogaxis}
\end{tikzpicture}
\caption{Median runtime as the number of updates increases. Both axes use logarithmic scales. Repeated single-record append exhibits cumulative quadratic behavior.}
\label{fig:runtime}
\end{figure}

\section{Application to Humanoid Robot Tasks}
\subsection{Public Benchmark-Derived Case Study}
To test interoperability with an independently developed task representation, we used the public BEHAVIOR-1K BDDL version 3.6.0 definition of the \texttt{storing\_food} activity~\cite{behavior1k}. BEHAVIOR-1K provides human-centered, long-horizon household activities as object sets, initial predicates, and logical goal conditions. We used the semantic task definition directly and did not run the OmniGibson physics simulator.

The initial condition places eight food objects---two oatmeal boxes, two bags of chips, two bottles of olive oil, and two jars of sugar---on a countertop. The goal requires every food object to be inside a cabinet. We mapped BDDL objects to \tws entities, binary predicates to \tws relations, and the task definition to a dataset observation. Every imported item records \texttt{source\_type=dataset}, the BDDL activity, package version, and a SHA-256 digest of the source task definition.

The conversion produced 13 entities and 12 initial relations. For each of the eight food objects, the semantic execution removes its \texttt{ontop} relation and adds an \texttt{inside} relation to a concrete cabinet instance, producing 16 ordered updates. The updates were written as 16 hash-linked event records and replayed from revision 0 to revision 16. All eight required goal predicates were satisfied after replay, and an independent invocation of the \tws command-line validator accepted the complete event log. Table~\ref{tab:behaviorcase} summarizes the result.

\begin{table}[t]
\caption{BEHAVIOR-1K storing-food case study.}\label{tab:behaviorcase}
\centering\small
\begin{tabular}{lr}\toprule
Measure & Result \\\midrule
BDDL version & 3.6.0 \\
Initial entities & 13 \\
Initial relations & 12 \\
Food objects & 8 \\
Hash-linked event records & 16 \\
Final revision & 16 \\
Goal predicates satisfied & 8/8 \\
Independent CLI verification & Pass \\\bottomrule
\end{tabular}
\end{table}

This case study evaluates semantic translation, provenance retention, sequence integrity, and deterministic replay. It does not evaluate physical feasibility, perception accuracy, motion planning, or humanoid control.

A humanoid manipulation system can use \tws between perception and
task-level decision making. For example, a vision component may create
an entity representing a container and attach an observation reference
and confidence score. A pose-estimation component may later replace
the entity with a refined estimate. A task component may assert a
relation such as an object being inside the container. Each accepted
transition receives a revision, permitting a planner decision to record
the exact state revision it consumed.

After a task failure, the event log can reconstruct the state at every
decision boundary. Provenance references identify which observations
supported an incorrect entity or relation, while the hash chain can
reveal whether the retained state history was subsequently modified.

This integration does not require \tws to transport images, point
clouds, or actuator data. Large sensor artifacts can remain in
specialized stores, with observations retaining identifiers and
metadata required to locate them. Adapters may translate between
\tws documents and ROS 2 messages without changing the core model.

\subsection{Cross-Task Import Evaluation}
We further evaluated whether the \tws representation could import semantic state from multiple independently authored BEHAVIOR-1K activities. The BDDL 3.6.0 installation contained 1,016 \texttt{problem0.bddl} task definitions. All 1,016 contained at least one grounded binary predicate in the initial-state section and therefore met our stated import eligibility criterion.

To avoid manually selecting favorable examples, we ranked eligible task paths by their SHA-256 digest and selected the first ten. The sample covered food preparation, household cleaning, storage, assembly, garment care, and labeling activities. For each task, BDDL objects referenced by grounded binary initial predicates were mapped to \tws entities, while those predicates were mapped to \tws relations. Each snapshot retained the BDDL version, activity name, source path, and SHA-256 digest of its source definition.

All ten sampled snapshots passed runtime validation and a subsequent invocation of the \tws command-line validator. In aggregate, the conversion produced 153 entities and 146 relations. Entity counts ranged from 5 to 46 per task, with a median of 11.5, while relation counts ranged from 4 to 45, with a median of 11. Table~\ref{tab:crosstask} summarizes the experiment.

\begin{table}[t]
\caption{Cross-task BEHAVIOR-1K import evaluation.}\label{tab:crosstask}
\centering\small
\begin{tabular}{lr}\toprule
Measure & Result \\\midrule
Available \texttt{problem0} tasks & 1,016 \\
Eligible task definitions & 1,016 \\
Deterministic sample size & 10 \\
Runtime validation & 10/10 \\
CLI validation & 10/10 \\
Total imported entities & 153 \\
Total imported relations & 146 \\
Entities per task (min/median/max) & 5/11.5/46 \\
Relations per task (min/median/max) & 4/11/45 \\\bottomrule
\end{tabular}
\end{table}

This experiment evaluates cross-task representational compatibility rather than complete BDDL semantics. It imports grounded binary initial-state predicates but does not execute quantified goals, simulate physical interaction, or measure task success on a robot.

\subsection{Public NVIDIA Humanoid Trajectories}
We evaluated \tws using the public NVIDIA \texttt{PhysicalAI-Robotics-GR00T-X-Embodiment-Sim} dataset~\cite{nvidiadata}, associated with the GR00T program~\cite{grootn1}. We pinned repository revision \path{ea7ac0b68f87da62f1e726771bba0fe74300802f} and used the \path{unitree_g1.LMPnPAppleToPlateDC} subset. The downloaded revision contained 103 episode-level Parquet files and 78,369 frames of simulated Unitree G1 state and action data at 50 Hz.

All source frames were parsed and checked for finite numeric values, consistent vector dimensions, monotonically increasing timestamps, and monotonically increasing frame indices. All 103 episodes passed these checks. For bounded event-log evaluation, 50 uniformly spaced frames, including the first and final frames, were recorded per episode. Each source Parquet file was linked to a \tws observation by repository revision and SHA-256 digest. The experiment produced 5,150 update records.

We compared \tws with a serialization-only Plain JSONL baseline containing the same update payloads. The baseline checked JSON parseability but provided no schema, sequence, provenance, or integrity validation. Replaying \tws reconstructed the complete mapped terminal row exactly for all 103 episodes; Plain JSONL also retained the terminal payload in all clean trials.

For each episode, we independently modified one record, deleted one middle record, exchanged two adjacent records, and duplicated one record. \tws detected all 103 trials of each corruption type, for 412/412 detections overall. The parse-only Plain JSONL baseline accepted all corrupted logs and therefore detected 0/412. This comparison isolates the checks absent from a serialization-only baseline; it does not establish superiority over integrity-protected logging systems. These tests evaluate committed middle-record corruption; they do not establish detection of suffix truncation without an externally trusted chain head or expected record count.

Across the 103 episode logs, Plain JSONL occupied 65.10 MB and \tws occupied 66.22 MB, corresponding to 1.72\% storage overhead. Median (IQR) time per 50-update episode was 6.284 (0.148) ms for Plain JSONL writing, 894.165 (5.660) ms for \tws append, 26.893 (0.247) ms for complete \tws verification, and 55.731 (0.709) ms for \tws replay. Plain JSONL terminal retrieval required 4.741 (0.234) ms. The append cost reflects the reference implementation's repeated whole-chain validation and is an optimization target rather than a constant-time streaming result. Table~\ref{tab:nvidia} summarizes the experiment.

\begin{table}[t]
\caption{NVIDIA Unitree G1 trajectory evaluation.}\label{tab:nvidia}
\centering\small
\begin{tabular}{lr}\toprule
Measure & Result \\\midrule
Episodes processed & 103/103 \\
Source frames checked & 78,369 \\
\tws updates recorded & 5,150 \\
Finite and consistent episodes & 103/103 \\
Terminal-state matches & 103/103 \\
\tws corruption detection & 412/412 \\
Plain JSONL corruption detection & 0/412 \\
Plain JSONL storage & 65.10 MB \\
\tws storage & 66.22 MB \\
\tws storage overhead & 1.72\% \\
Median \tws append per episode & 894.165 ms \\
Median \tws verification & 26.893 ms \\
Median \tws replay & 55.731 ms \\\bottomrule
\end{tabular}
\end{table}

\subsection{Artifact Availability}
The core runtime is maintained at \url{https://github.com/xli431/tws}. The inspected revision \path{cedcbdc9ba8d732f777466b41c7621d797836800} contains the implementation, schemas, examples, and conformance tests. At preparation of this preprint the repository is private. The inspected revision does not contain the runtime benchmark harness or the BEHAVIOR-1K and NVIDIA experiment scripts and raw results; consequently, it is not yet a complete reproduction artifact for the reported evaluations.

\section{Limitations}

The current evaluation uses a controlled reference lifecycle, semantic task definitions, and simulated humanoid trajectories rather than a physical humanoid deployment. It therefore validates state and
log behavior, not perception accuracy, control performance, or task
success. The representation also does not yet define coordinate-frame
transforms, conflict-resolution policies, ontology alignment, or
probabilistic fusion.

A SHA-256 chain alone cannot authenticate the writer or protect
against complete history replacement. Version 0.1 requires an
external trusted digest for that threat model. Finally, the storage
implementation assumes a single writer and does not yet specify
recovery from an interrupted final record.

These limitations define concrete next steps: integration with a
humanoid manipulation pipeline, task-scale performance evaluation,
trusted checkpoints or signatures, explicit coordinate-frame
semantics, and safe concurrent storage.

\section{Conclusion}

This paper presented \tws, a provenance-aware representation and
reference runtime for traceable robot world state. \tws combines
versioned snapshots, explicit evidence references, confidence,
validated immutable updates, deterministic replay, and a
tamper-evident event log. The prototype passes 38 conformance tests
across three Python versions and detects the evaluated transition and
log-integrity faults.

The design is intentionally independent of robot middleware and
hardware. This makes it suitable as a compact experiment contract for
humanoid perception and manipulation systems, where reconstructing
what the robot believed, why it believed it, and how that state changed
is essential for debugging and reproducible evaluation.

\end{document}